\documentclass[10pt,twocolumn,letterpaper]{article}
\usepackage{cvpr}
\usepackage{times}
\usepackage{graphicx}
\usepackage{amsmath}
\usepackage{amssymb}
\usepackage{xspace}
\usepackage{xcolor}
\usepackage{enumitem}
\usepackage{listings}
\usepackage{tabularx}
\usepackage{multirow}
\usepackage{gensymb}
\usepackage{svg}
\usepackage{pgfplots}
\usepackage{pgfkeys}
\usepackage{adjustbox}
\usepackage{array}
\usepackage{multirow}
\usepackage{gensymb}
\usepackage[font=small,labelfont=bf]{caption} % Just test...
\usepackage[numbers,square,sort&compress]{natbib}
\usepackage{booktabs}% http://ctan.org/pkg/booktabs
\usepackage{xparse}% http://ctan.org/pkg/xparse
\usepackage[
  pagebackref=true,
  breaklinks=true,
  letterpaper=true,
  colorlinks,
  bookmarks=false]{hyperref}

\cvprfinalcopy
\def\cvprPaperID{530}
\ifcvprfinal\fi

\graphicspath{{./}}

\setitemize{noitemsep,topsep=0pt,parsep=0pt,partopsep=0pt,nolistsep,leftmargin=*}

\renewcommand{\paragraph}[1]{\medskip\noindent{\bf #1}}

\definecolor{attr}{rgb}{0.85,0.88,0.90}
\definecolor{code}{rgb}{0.95,0.93,0.90}

\newcommand{\off}[1]{}

\newcommand{\bb}{\mathbf{b}}
\newcommand{\bx}{\mathbf{x}}
\newcommand{\by}{\mathbf{y}}
\newcommand{\bz}{\mathbf{z}}
\newcommand{\bw}{\mathbf{w}}

\newcommand{\real}{\mathbb{R}}

\pgfplotsset{compat=newest}
\pgfplotsset{
	tick label style={font=\footnotesize},
	label style={font=\footnotesize},
	legend style={font=\footnotesize},
	title style={font=\footnotesize}}
\newlength\figureheight
\newlength\figurewidth

\NewDocumentCommand{\rot}{O{45} O{1em} m}{\makebox[#2][c]{\rotatebox{#1}{#3}}}%

\makeatletter
\g@addto@macro\normalsize{%
  \setlength\abovedisplayskip{0.7em}
  \setlength\belowdisplayskip{0.7em}
  \setlength\abovedisplayshortskip{0.7em}
  \setlength\belowdisplayshortskip{0.7em}
}
\makeatother

\newenvironment{customlegend}[1][]{%
   	\begingroup
   	\csname pgfplots@init@cleared@structures\endcsname
   	\pgfplotsset{#1}%
}{%
\csname pgfplots@createlegend\endcsname
\endgroup
}%

\def\addlegendimage{\csname pgfplots@addlegendimage\endcsname}

\begin{document}
\title{Understanding image representations \\
by measuring their equivariance and equivalence}
\vspace{-2em}
%\author{
%Karel Lenc
%~~~~~~~~~~~~~~~~~~~ %\and
%Andrea Vedaldi \\
%	Visual Geometry Group, University of Oxford\\
%	\texttt{\{karel;vedaldi\}@robots.ox.ac.uk}}
\author{Karel Lenc ~~~~~~~~~~~~~~~~~~ Andrea Vedaldi\\
	\small Department of Engineering Science, University of Oxford\\
	%{\tt\small \{karel,vedaldi\}@robots.ox.ac.uk}
	%\and
}
\maketitle
% --------------------------------------------------------------------

% TODO!
%R3:. The paper does not analyze the learnt transformations Mg
%There are several conclusions in the paper that we will endeavour to better emphasise in the final version:

%(1) The key result is that the CNN features globally change in an easily predictable way in term of linear transformations; importantly, the *same* linear transformation works for *any* input image, and hence *any* object category, suggesting that geometry is factored in a uniform way for all of them;

%(2) (l.644, Fig.6) while the representation is distributed, information transforms locally, both spatially and in term of feature channels; this can be seen because the support of the transformation filter is small, these filters are sparse, and using only the closest lattice sites (l.359) is sufficient for prediction;

%(3) We found and quantified the number of invariant feature components in each layer of the representation.

% --------------------------------------------------------------------
\begin{abstract}
Despite the importance of image representations such as histograms of oriented gradients and deep Convolutional Neural Networks (CNN), our theoretical understanding of them remains limited. Aiming at filling this gap, we investigate three key mathematical properties of representations: equivariance, invariance, and equivalence. Equivariance studies how transformations of the input image are encoded by the representation, invariance being a special case where a transformation has no effect. Equivalence studies whether two representations, for example two different parametrisations of a CNN, capture the same visual information or not. A number of methods to establish these properties empirically are proposed, including introducing transformation and stitching layers in CNNs. These methods are then applied to popular representations to reveal insightful aspects of their structure, including clarifying at which layers in a CNN certain geometric invariances are achieved. While the focus of the paper is theoretical, direct applications to structured-output regression are demonstrated too.
\end{abstract}
% --------------------------------------------------------------------

%While image representations are widely used in computer vision, their theoretical understanding remains very limited. Generally sought representational properties include \emph{discriminability} of factors of interest, such as an object class, and \emph{invariance} to nuisance factors, such as viewpoint; alternatively, representations are thought to \emph{untangle} independent factors~\cite{bengio13representation}. However, these characterisations are rather vague and it remains unclear which information is captured by representations, which invariances they posses, and how these properties are obtained.

% --------------------------------------------------------------------
\section{Introduction}\label{s:intro}
% --------------------------------------------------------------------

Image representations have been a key focus of the research in computer vision for at least two decades. Notable examples include textons~\cite{leung01representing}, histogram of oriented gradients (SIFT~\cite{lowe04distinctive} and HOG~\cite{dalal05histograms}), bag of visual words~\cite{csurka04visual}\cite{sivic03video}, sparse~\cite{yang10supervised} and local coding~\cite{wang10locality-constrained}, super vector coding~\cite{zhou10image}, VLAD~\cite{jegou10aggregating}, Fisher Vectors~\cite{perronnin06fisher}, and the latest generation of deep convolutional networks~\cite{krizhevsky12imagenet,ZeilerArxiv13,sermanet14overfeat:}. However, despite their popularity, our theoretical understanding of representations remains limited. It is generally believed that a good representation should combine invariance and discriminability, but this characterisation is rather vague; for example, it is often unclear what invariances are contained in a representation and how they are obtained.

\begin{figure}[ht!]
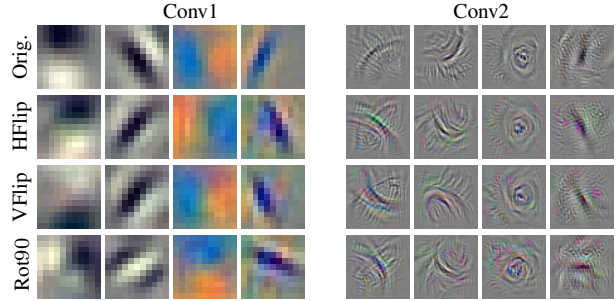

	\vspace{-0.4em}
	\begin{center}
		\begin{footnotesize}
			\setlength{\tabcolsep}{1pt}
			\newcommand{\CFImC}[3]{\includegraphics[width=0.1\linewidth]{figures/conv#1_tf_fs/#2/#3.png}}
			\begin{tabular}{ r c c c c @{\hskip 2em} c c c c}
				        &  \multicolumn{4}{c }{Conv1} & \multicolumn{4}{c}{Conv2} \\
				\rot[90]{\hspace{0.2em} Orig.}   & 
				\CFImC{1}{orig}{045}         & \CFImC{1}{orig}{005}        & \CFImC{1}{orig}{088}       & \CFImC{1}{orig}{050}        & 
				\CFImC{2}{orig}{002}         & \CFImC{2}{orig}{048}        & \CFImC{2}{orig}{022}       & \CFImC{2}{orig}{061}        \\
				\rot[90]{\hspace{0.1em} HFlip} &
				\CFImC{1}{horiz-flip}{045}   & \CFImC{1}{horiz-flip}{005}  & \CFImC{1}{horiz-flip}{088} & \CFImC{1}{horiz-flip}{050}        & 
				\CFImC{2}{horiz-flip}{002}   & \CFImC{2}{horiz-flip}{048}  & \CFImC{2}{horiz-flip}{022} & \CFImC{2}{horiz-flip}{061}        \\ 
				\rot[90]{\hspace{0.07em} VFlip} &
				\CFImC{1}{vert-flip}{045}    & \CFImC{1}{vert-flip}{005}   & \CFImC{1}{vert-flip}{088}  & \CFImC{1}{vert-flip}{050}        & 
				\CFImC{2}{vert-flip}{002}    & \CFImC{2}{vert-flip}{048}   & \CFImC{2}{vert-flip}{022}  & \CFImC{2}{vert-flip}{061}        \\ 
				\rot[90]{\hspace{0.1em} Rot$90$} &
				\CFImC{1}{rot-90}{045}       & \CFImC{1}{rot-90}{005}      & \CFImC{1}{rot-90}{088}     & \CFImC{1}{rot-90}{050}        & 
				\CFImC{2}{rot-90}{002}       & \CFImC{2}{rot-90}{048}      & \CFImC{2}{rot-90}{022}     & \CFImC{2}{rot-90}{061}        \\ 
			\end{tabular}
			\vspace{-1em}
			\caption{{\bf Equivariant transformation of CNN filters.} Top: Conv1 and Conv2 filters of a convolutional neural network visualised with the method of~\cite{simonyan2013deep}. Other rows: geometrically warped filters reconstructed from an equivariant transformation of the network output learned using the method of Sect.~\ref{s:learning} for Horizontal flip, Vertical flip and Rotation $90\degree$.}
			\label{fig:filtvis}
		\end{footnotesize}
	\end{center}
\end{figure}

In this work, we propose a new approach to study image representations. We look at a \emph{representation} $\phi$ as an abstract function mapping an image $\bx$ to a vector $\phi(\bx)\in\real^d$ and we \emph{empirically establish key mathematical properties} of this function. We focus in particular on three such properties (Sect.~\ref{s:learning}). The first one is {\bf equivariance}, which looks at how the representation \emph{changes upon transformations of the input image}. We demonstrate that most representations, including HOG and most of the layers in deep neural networks, change in a \emph{easily predictable} manner with the input (Fig.~\ref{fig:filtvis}). We show that such equivariant transformations can be learned empirically from data (Sect.~\ref{s:regression}) and that, importantly, they amount to simple linear transformations of the representation output (Sect.~\ref{s:experiments-shallow} and~\ref{s:experiments-deep}). In the case of convolutional networks, we obtain this by introducing and learning a new \emph{transformation layer}. By analysing the learned equivariant transformations we are also able to find and characterise the {\bf invariances} of the representation, our second property. This allows us to quantify invariance and show how it builds up with depth in deep models.

The third property, {\bf equivalence}, looks at whether the information captured by heterogeneous representations is in fact the same. CNN models, in particular, contain millions of redundant parameters~\cite{denil13predicting} that, due to non-convex optimisation in learning, may differ even when retrained on the same data. The question then is whether the resulting differences are genuine or just apparent. To answer this question we learn \emph{stitching layers} that allow swapping parts of different networks. Equivalence is then obtained if the resulting ``Franken-CNNs'' perform as well as the original ones (Sect.~\ref{s:experiments-equivalence}).

The rest of the paper is organised as follows. Sect.~\ref{s:learning} discussed methods to learn empirically representation equivariance, invariance, and equivalence. Sect.~\ref{s:experiments-shallow} and~\ref{s:experiments-deep} present experiments on shallow and deep representation equivariance respectively, and Sect.~\ref{s:experiments-equivalence} on representation equivalence. Sect.~\ref{s:so} demonstrates a practical application of equivariant representations to structured-output regression. Finally, Sect.~\ref{s:summary} summarises our findings.

\paragraph{Related work.} The problem of designing invariant or equivariant features has been widely explored in computer vision. For example, a popular strategy is to extract invariant local descriptors~\cite{lowe99object} on top of equivariant (also called co-variant) detectors~\cite{lindeberg98principles,mikolajc03survey,lowe99object}. Various authors have also looked at incorporating equivariance explicitly in the representations~\cite{schimdt12learning, sohn2012learning}. Deep CNNs, including the one of Krizhevsky~\etal~\cite{krizhevsky12imagenet} and related state-of-the-art architecutres, are deemed to build an increasing amount of invariance layer after layer. This is even more explicit in the \emph{scattering transform} of Sifre and Mallat~\cite{sifre13rotation}.

In all these examples, invariance is a design aim that may or may not be achieved by a given architecture. By contrast, our aim is \emph{not} to propose yet another mechanism to learn invariances, but rather a method to systematically tease out invariance, equivariance, and other properties that a given representation may have. To the best of our knowledge, there is very limited work in conducting this type of analysis. Perhaps the contributions that come closer study only invariances of neural networks to specific image transformations~\cite{goodfellow09,ZeilerArxiv13}. However, we believe to be the first to functionally characterise and quantify these properties in a systematic manner, as well as being the first to investigate the equivalence of different representations.

%TODO: fastest pedestrian

% --------------------------------------------------------------------
\section{Notable properties of representations}\label{s:learning}
% --------------------------------------------------------------------

\emph{Image representations} such as HOG, SIFT, or CNNs can be thought of as functions $\phi$ mapping an image $\bx\in\mathcal{X}$ to a vector $\phi(\bx)\in\real^d$. This section describes three notable properties of representations --- equivariance, invariance, and equivalence --- and gives algorithms to establish them empirically.

\paragraph{Equivariance.} A representation $\phi$ is \emph{equivariant} with a transformation $g$ of the input image if the transformation can be transferred to the representation output. Formally, equivariance with $g$ is obtained when there exists a map $M_g : \real^d \rightarrow \real^d$ such that:
\begin{equation}\label{e:equiv}
   \forall \bx \in \mathcal{X} : \quad \phi(g\bx) \approx M_g \phi(\bx).
\end{equation}
A \emph{sufficient condition} for the existence of $M_g$ is that the representation $\phi$ is \emph{invertible}, because in this case $M_g = \phi \circ g \circ \phi^{-1}$. It is known that representations such as HOG are at least approximately invertible~\cite{vondrick13hoggles:}. Hence it is not just the existence, but also the structure of the mapping $M_g$ that is of interest. In particular, $M_g$ should be \emph{simple}, for example a linear function. This is important because the representation is often used in simple predictors such as linear classifiers, or in the case of CNNs, is further processed by linear filters. Furthermore, by requiring the \emph{same} mapping $M_g$ to work for \emph{any} input image, intrinsic geometric properties of the representations are captured.

The nature of the transformation $g$ is in principle arbitrary; in practice, in this paper we will focus on geometric transformations such as affine warps and flips of the image.

\paragraph{Invariance.} Invariance is a special case of equivariance obtained when $M_g$ (or a subset of $M_g$) acts as the simplest possible transformation, i.e. the identity map. Invariance is often regarded as a key property of representations since one of the goals of computer vision is to establish invariant properties of images. For example, the category of the objects contained in an image is invariant to viewpoint changes. By studying invariance systematically, it is possible to clarify if and where the representation achieves it.

%\footnote{Exact invariance, requires closure with respect to the transformation group~\cite{vedaldi05features}: if $\phi(\bx) = \phi(g \bx)$ and $\phi(g \bx) = \phi(g' g\bx)$, then $\phi(\bx) = \phi(gg'\bx)$. For most transformations, this is not realistic -- for example, if $t$ is a scaling, exact group closure means that invariance should hold even if the image is reduced to a single pixel.} 

\paragraph{Equivalence.} While equi/invariance look at how a representation is affected by transformations of the image, equivalence studies the relationship between different representations. Two heterogenous representations $\phi$ and $\phi'$ are \emph{equivalent} if there exist a map $E_{\phi \rightarrow \phi'}$ such that
\[
 \forall \bx: \quad  \phi'(\bx) \approx E_{\phi \rightarrow \phi'} \phi(\bx).
\]
If $\phi$ is invertible, then $E_{\phi \rightarrow \phi'} = \phi' \circ \phi^{-1}$ satisfies this condition; hence, as for the mapping $M_g$ before, the interest is not just in the existence but also in the structure of the mapping $E_{\phi \rightarrow \phi'}$.

\paragraph{Example: equivariant HOG transformations.} Let $\phi$ denote the HOG~\cite{dalal05histograms} feature extractor. In this case $\phi(\bx)$ can be interpreted as a $H \times W$ vector field of of $D$-dimensional feature vectors or cells. If $g$ denotes image flipping around the vertical axis, then $\phi(\bx)$ and $\phi(g\bx)$ are related by a well defined \emph{permutation} of the feature components. This permutation swaps the HOG cells in the horizontal direction and, within each HOG cell, swaps the components corresponding to symmetric orientations of the gradient. Hence the mapping $M_g$ is a permutation and one has \emph{exactly} $\phi(g\bx) = M_g \phi(\bx)$. The same is true for horizontal flips and $180\degree$ rotations, and, approximately,\footnote{Most HOG implementations use 9 orientation bins, breaking rotational symmetry.} for $90\degree$ rotations. HOG implementations~\cite{vedaldi10vlfeat} do in fact explicitly provide such permutations.

\paragraph{Example: translation equivariance in convolutional representations.} HOG, densely-computed SIFT (DSIFT), and convolutional networks are examples of \emph{convolutional representations} in the sense that they are obtained from local and translation invariant operators. Barring boundary and sampling effects, any convolutional representation is equivariant to translations of the input image as this result in a translation of the feature field.

% ----------------------------------------------------------
\subsection{Learning properties with structured sparsity}\label{s:regression}
% ----------------------------------------------------------

When studying equivariance and equivalence, the transformation $M_g$ and $E_{\phi\rightarrow\phi'}$ are usually not available in closed form and must be \emph{estimated} from data. This section discusses a number of algorithms to do so. The discussion focuses on equivariant transformations $M_g$, but dealing with equivalence transformations $E_{\phi\rightarrow\phi'}$ is similar.

Given a representation $\phi$ and a transformation $g$, the goal is to find a mapping $M_g$ satisfying~\eqref{e:equiv}. In the simplest case $M_g = (A_g, \bb_g),$ $A_g\in\real^{d\times d},$ $\bb_g\in\real^d$ is an affine transformation $\phi(g\bx) \approx A_g \phi(\bx) + \bb_g$. This choice is not as restrictive as it may initially seem: in the examples above $M_g$ is a permutation, and hence can be implemented by a corresponding permutation matrix $A_g$.

Estimating $(A_g,\bb_g)$ is naturally formulated as an empirical risk minimisation problem. Given data $\bx$ sampled from a set of natural images, learning amounts to optimising the regularised reconstruction error
\begin{equation}\label{e:objective}
E(A_g,\bb_g) = \lambda \mathcal{R}(A_g) + \frac{1}{n} \sum_{i=1}^n \ell(\phi(g\bx_i), A_g \phi(\bx_i) + \bb_g),
\end{equation}
where $\mathcal{R}$ is a regulariser and $\ell$ a regression loss whose choices are discussed below. The objective~\eqref{e:objective} can be adapted to the equivalence problem by replacing $\phi(g\bx)$ by $\phi'(\bx)$.

\paragraph{Regularisation.} The choice of regulariser is particularly important as $A_g \in \real^{d\times d}$ has a $\Omega(d^2)$ parameters. Since $d$ can be quite large (for example, in HOG one has $d = DWH$), regularisation is essential. The standard $l^2$ regulariser $\|A_g\|_F^2$ was found to be inadequate; instead, sparsity-inducting priors work much better for this problem as they encourage $A_g$ to be similar to a permutation matrix.

We consider two such sparsity-inducing regularisers. The first regulariser allows $A_g$ to contain a fixed number $k$ of non-zero entries for each row:
\begin{equation}
\mathcal{R}_k(A) = 
\begin{cases}
+\infty, & \exists i: \| A_{i,:} \|_0 > k, \\
\|A\|_F^2, & \text{otherwise}.
\end{cases}\label{e:ff}
\end{equation}
Regularising rows independently reflects the fact that each row is a predictor of a particular component of $\phi(g\bx)$.

\begin{figure}
\centering
\includegraphics[width=0.8\columnwidth]{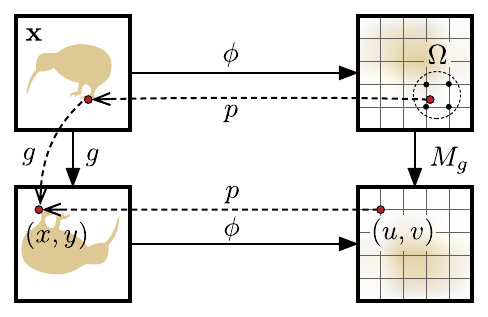}
\caption{{\bf Structured sparsity.} Predicting equivariant features at location $(u,v)$ uses a corresponding small neighbourhood of features $\Omega_{g,m}(u,v)$.}\label{f:neighbours}
\end{figure}

The second sparsity-inducing regulariser is similar, but exploits the \emph{convolutional} structure of many representations. Convolutional features are obtained from translation invariant and local operators (non-linear filters), such that the representation $\phi(\bx)$ can be interpreted as a feature field with spatial indexes $(u,v)$ and channel index $t$. Due to the locality of the representation, the component $(u,v,t)$ of $\phi(g\bx)$ should be predictable from a corresponding neighbourhood $\Omega_{g,m}(u,v)$ of features in the feature field $\phi(\bx)$ (Fig.~\ref{f:neighbours}). This results in a particular sparsity structure for $A_g$ that can be imposed by the regulariser
\begin{equation}\label{e:ss}
\mathcal{R}_{g,m}(A)=
\begin{cases}
+\infty, &
\parbox[t]{.23\textwidth}{\raggedright$\exists t, t', (u,v), (u',v') \not\in \Omega_{g,m}(u,v) : A_{uvt,u'v't'}\not=0$} \\
\|A\|_F^2, &\text{otherwise,} \\
\end{cases}
\end{equation}
where $m$ denotes the neighbour size and indexes of $A$ have been identified with triplets $(u,v,t)$. The neighbourhood itself is defined as the $m \times m$ input feature sites closer to the back-projection of the output feature $(u,v)$.\footnote{Formally, denote by $(x,y)$ the coordinates of a pixel in the input image $\bx$ and by $p:(u,v)\mapsto(x,y)$ the affine function mapping the feature index $(u,v)$ to the centre $(x,y)$ of the corresponding \emph{receptive field} (measurement region) in the input image. Denote by $\mathcal{N}_k(u,v)$ the $k$ feature sites $(u',v')$ that are closer to $(u,v)$ (the latter can have fractional coordinates) and use this to define the neighbourhood of the back-transformed site $(u,v)$ as $\Omega_{g,k}(u,v) = \mathcal{N}_k(p^{-1}\circ g^{-1} \circ p(u,v))$.} In practice \eqref{e:ff} and \eqref{e:ss} will be combined in order to limit the number of regression coefficients activated in each neighbourhood. 

\paragraph{Loss.} As empirically shown in Sect.~\ref{s:experiments-deep}, the choice of loss $\ell$ is important. For HOG and similar histogram-like features, the $l^2$, Hellinger's, or $\chi^2$ distances work well. However, for more sophisticated features such as deep layers in CNNs, it was found that target-oriented losses can perform substantially better in certain cases. To understand the concept of target-oriented loss, consider a CNN $\phi$ trained end-to-end on a categorisation problem such as the ILSVRC 2012 image classification task~(ILSVRC12)~\cite{ILSVRCarxiv14}. A common approach \cite{Razavian2014, Donahue2013, chatfield14return} is to use the first several layers $\phi_1$ of $\phi = \phi_2 \circ \phi_1$ as a general-purpose feature extractor. This suggests an alternative objective that preserves the quality of the equivariant features $\phi_1$ in the original problem:
\begin{multline}\label{e:objective2}
E(A_g,\bb_g) = \lambda \mathcal{R}(A_g) + \\
\frac{1}{n} \sum_{i=1}^n \ell(y_i, 
\phi_2 \circ (A_g,\bb_g) \circ \phi_1(g^{-1} \bx_i)).
\end{multline}
Here $y_i$ denotes the ground truth label of image $\bx_i$ and $\ell$ is the same classification loss used to train $\phi$. Note that in this case $(A_g,\bb_g)$ is learned to \emph{compensate} for the image transformation, which therefore is set to $g^{-1}$. This formulation is not restricted to CNNs, but applies to any representation $\phi_1$ given a target classification or regression task and a corresponding pre-trained predictor $\phi_2$ for it.

% ----------------------------------------------------------
\subsection{Equivariance in CNNs: transformation layers}\label{s:transformation}
% ----------------------------------------------------------

The method of Sect.~\ref{s:regression} can be substantially refined for the case of convolutional representations and certain transformation classes. The structured sparsity regulariser~\eqref{e:ss} encourages $A_g$ to match the convolutional structure of the representation. If $g$ is an affine transformation more can be said: up to sampling artefacts, the equivariant transformation $M_g$ is local and translation invariant, \ie convolutional. The reason is that an affine $g$ acts uniformly on the image domain\footnote{This means that $g(x+u,y+v) = g(x,y) + (u',v')$.} so that the same is true for $M_g$. This has two key advantages: it reduces dramatically the number of parameters to learn and it can be implemented efficiently as an additional layer of a CNN. Such a \emph{transformation layer} consists of a \emph{permutation layer} that maps input feature sites $(u,v,t)$ to output feature sites $(g(u,v),t)$ followed by a bank of $D$ linear filters, each of dimension $m \times m \times D$. Here $m$ corresponds to the size of the neighbourhood $\Omega_{g,m}(u,v)$ in Sect.~\ref{s:regression}. Intuitively, the main purpose of these filters is to permute and interpolate feature channels.

Note that $g(u,v)$ does not, in general, fall at integer coordinates.  In our case, the permutation layer assigns $g(u,v)$ to the closest lattice site by rounding but it can be also distributed to the nearest $2\times 2$ sites by using bilinear interpolation.\footnote{Better accuracy could be obtained by using image warping techniques. For example, sub-pixel accuracy can be obtained by upsampling in the permutation layer and then allowing the transformation filter to be translation variant (or, equivalently, by introducing a suitable non-linear mapping between the permutation layer and transformation filters).}

% ----------------------------------------------------------
\subsection{Equivalence in CNNs: stitching layers}\label{s:stitching}
% ----------------------------------------------------------

The previous section looked at how equivariance can be studied more efficiently in CNNs; this section does the same for equivalence. Following the task-oriented loss formulation of Sect.~\ref{s:regression}, consider two representations $\phi_1$ and $\phi_1'$ and a predictor $\phi_2'$ learned to solve a reference task using the representation $\phi_1'$. For example, these could be obtained by decomposing two CNNs $\phi = \phi_2 \circ \phi_1$ and $\phi' = \phi_2' \circ \phi_1'$ trained on the ImageNet ILSVCR data (but $\phi_1$ could also be learned on a different problem or be handcrafted).

The goal is to find a mapping $E_{\phi_1\rightarrow\phi_1'}$ such that $\phi_1' \approx E_{\phi_1\rightarrow\phi_1'} \phi_1$. This map can be seen as a ``stitching transformation'' allowing $\phi_2' \circ E_{\phi_1\rightarrow\phi_1'} \circ \phi_1$ to perform as well as $\phi_2' \circ \phi_1'$ on the original classification task. Hence this transformation can be learned by minimizing the loss $\ell(y_i, \phi_2' \circ E_{\phi_1\rightarrow\phi_1'} \circ \phi_1(\bx_i))$ in an objective similar to~\eqref{e:objective2}. In a CNN the map $E_{\phi_1\rightarrow\phi_1'}$ can be interpreted as a \emph{stitching layer}. Furthermore, given the convolutional structure of the representation, this layer can be implemented as a bank of linear filters. No permutation layer is needed in this case, but it may be necessary to down/upsample the features if the spatial dimensions of $\phi_1$ and $\phi_1'$ do not match.

\section{Experiments}\label{s:experiments}
% ----------------------------------------------------------

The experiments begin in Sect.~\ref{s:experiments-shallow} by studying the problem of learning equivariant mappings for shallow representations. Sect.~\ref{s:experiments-deep} and \ref{s:experiments-equivalence} move on to deep convolutional representations, examining equivariance and equivalence respectively.
In Sect.~\ref{s:so} equivariant mappings are applied to structure-output regression.
% ----------------------------------------------------------
\subsection{Equivariance in shallow representations}\label{s:experiments-shallow}
% ----------------------------------------------------------

\begin{figure*}
\centering
\vspace*{-4em}
\makebox[\textwidth][l]{
\hspace*{-1.5em}
\captionsetup[subfigure]{oneside,margin={2.3em,0em}}
\subfloat[{Rotation $[^{\circ}]$}]{
	\setlength{\figureheight}{2.3cm}
	\setlength{\figurewidth}{0.22\linewidth}
	\input{figures/hog_dense_optres_rot.tkiz}
	\label{fig:hog_mg_rot}}

\subfloat[{Iso-scale $2^x$}]{
	\setlength{\figureheight}{2.3cm}
	\setlength{\figurewidth}{0.22\linewidth}
	\input{figures/hog_dense_optres_sc.tkiz}
	\label{fig:hog_mg_sc}}
\captionsetup[subfigure]{oneside,margin={2em,0em}}
\subfloat[{Neighbourhood size $m$ [cells] }]{
	\setlength{\figureheight}{2.3cm}
	\setlength{\figurewidth}{0.22\linewidth}
	\input{figures/hog_gprior_optres.tkiz}
	\label{fig:hog_mg_cell}}
%\hspace*{1em}
\subfloat{
\begin{tikzpicture}
\definecolor{mycolor1}{rgb}{0.10588,0.61961,0.46667}%
\definecolor{mycolor2}{rgb}{0.85098,0.37255,0.00784}%
\definecolor{mycolor3}{rgb}{0.45882,0.43922,0.70196}%
\begin{customlegend}[legend columns=1,
		legend style={align=left,draw=none},
		legend cell align=left,
		legend entries={None, LS, RR $\lambda = 1$, RR $\lambda = 0.1$, FS $k = 1$, FS $k = 5$, FS $k = 25$}]
	\addlegendimage{color=mycolor1,solid,line width=1.0pt,mark size=1.5pt,mark=+,mark options={solid}}
	\addlegendimage{color=mycolor2,solid,line width=1.0pt,mark size=1.5pt,mark=+,mark options={solid}}   
	\addlegendimage{color=mycolor3,solid,line width=1.0pt,mark size=1.5pt,mark=+,mark options={solid}}
	\addlegendimage{color=mycolor3,dash pattern=on 1pt off 3pt on 3pt off 3pt,line width=1.0pt,mark size=1.5pt,mark=+,mark options={solid}}
	\addlegendimage{color=white!40!black,dotted,line width=1.0pt,mark size=1.5pt,mark=+,mark options={solid}}
	\addlegendimage{color=white!40!black,dashed,line width=1.0pt,mark size=1.5pt,mark=+,mark options={solid}}
	\addlegendimage{color=white!40!black,solid,line width=1.0pt,mark size=1.5pt,mark=+,mark options={solid}}
\end{customlegend}
\end{tikzpicture}
	\label{fig:hog_mg_legend}}
}
\vspace{-1em}
\caption{{\bf Regression methods.} The figure reports the HOG feature reconstruction error (average per-cell Hellinger distance) achieved by the learned equivariant mapping $M_g$ by setting $g$ to different image rotations (\ref{fig:hog_mg_rot}) and scalings (\ref{fig:hog_mg_sc}) for different learning strategies (see text). No other constraint is imposed on $A_g$. In the right panel (\ref{fig:hog_mg_cell}) the experiment is repeated for the $45^\circ$ rotation, but this time imposing structured sparsity on $A_g$ for different values of the neighbourhood size $m$.}\label{f:regression}
\end{figure*}

\begin{figure}
	\begin{center}
		\setlength{\figureheight}{2.5cm}
		\setlength{\figurewidth}{0.4\linewidth}
		\hspace*{-0.2in}
		\input{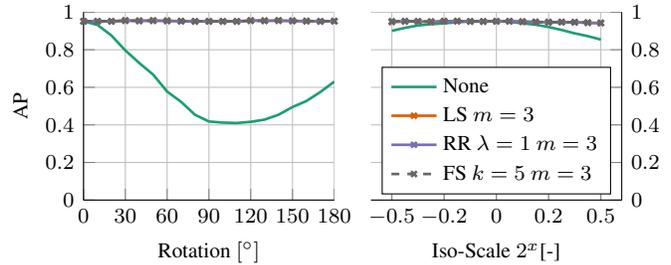}
	\end{center}
	\vspace{-2em}
	\caption{{\bf Equivariant classification using HOG features.}  Classification performance of a HOG-based classifier trained to discriminate dog and cat heads as the test images are gradually rotated and scaled and the effect compensated by equivariant maps learned using LS, RR, and FS.}
	\label{fig:gprior_vs_dist}
	\label{fig:cathead_rot_sc_exp}
\end{figure}

\begin{table}
	\vspace{-1em}
\begin{center}
\begin{footnotesize}\begin{tabular}{|cc|r|r|r|r|}
\hline
& & \multicolumn{4}{c|}{HOG size} \\
$k$ &$m$ &\textbf{$3 \times 3$}&\textbf{$5 \times 5$}&\textbf{$7 \times 7$}&\textbf{$9 \times 9$}\\\hline
5&$\infty$&1.67&12.21&82.49&281.18\\
5&1&0.97&2.06&3.47&5.91\\
5&3&1.23&3.90&7.81&13.04\\
5&5&1.83&7.46&17.96&30.93\\
\hline\end{tabular}
\end{footnotesize}
\vspace{-0.5em}
\caption{{\bf Regression cost.} Cost (in seconds) of learning the equivariant regressors of Fig.~\ref{fig:gprior_vs_dist}. As the size of the HOG arrays becomes larger, the optimisation cost increases significantly unless structured sparsity is considered by setting $m$ to a small number.}
\label{tab:gprior-time}
\end{center}
\end{table}

\begin{figure}
\vspace{-0.2em}
\begin{center}
\begin{footnotesize}
\setlength{\tabcolsep}{1pt}
\newcommand{\HglInsIm}[1]{\includegraphics[width=0.18\linewidth]{figures/hoggles/#1}}
\newcommand{\HglRotIm}[1]{\HglInsIm{n02111889_5259/rot-45/#1}}
\newcommand{\HglDownScIm}[1]{\HglInsIm{n11939491_19368/scale--0.50/#1}}
\newcommand{\HglUpScIm}[1]{\HglInsIm{n03476991_48224/scale-0.50/#1}}
\begin{tabular}{ r c c c c}
                                         & $\bx$                    & $\phi^{-1}\phi(\bx)$             & $\phi^{-1}\phi(g\bx)$          & $\phi^{-1} M_g \phi(\bx)$          \\
\rot[90][2em]{\hspace{0.3em} $g=$Rot $45^{\degree}$}    & \HglRotIm{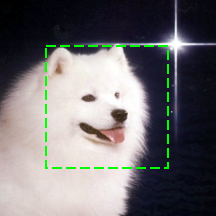} & \HglRotIm{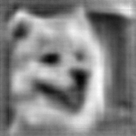}    & \HglRotIm{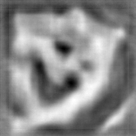}    & \HglRotIm{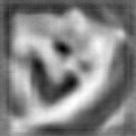}    \\
\rot[90][2em]{\hspace{0.2em} $g=$Sc $2^{-\frac{1}{2}}$} & \HglDownScIm{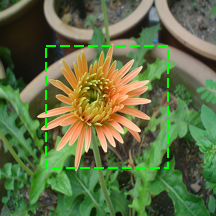} & \HglDownScIm{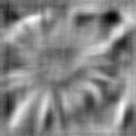} & \HglDownScIm{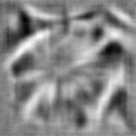} & \HglDownScIm{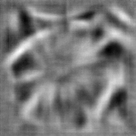} \\
\rot[90][2em]{\hspace{0.4em} $g=$Sc $2^{\frac{1}{2}}$}  & \HglUpScIm{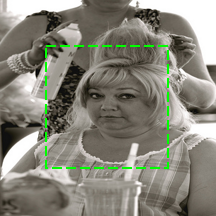}   & \HglUpScIm{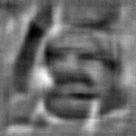}   & \HglUpScIm{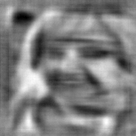}   & \HglUpScIm{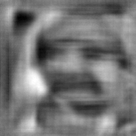}
\end{tabular}
\vspace{-1em}
\caption{{\bf Qualitative evaluation of equivariant HOG.} Visualisation of the features $\phi(\bx)$, $\phi(g\bx)$ and $M_g\phi(\bx)$ using the $\phi^{-1}$ HOGgle~\cite{vondrick13hoggles:} HOG inverse. $M_g$ is learned using FS with $k=5$ and $m=3$ and $g$ is set to a rotation by $45\degree$ and up/down-scaling by $\sqrt{2}$ respectively. The dashed boxes show the support of the reconstructed features.}
\label{fig:hoggles}
\end{footnotesize}
\end{center}
\vspace{-1em}
\end{figure}
This section applies the methods of Sect.~\ref{s:regression} to learn equivariant maps for shallow representations, and HOG features in particular. The first method to be evaluated is sparse regression, followed  by structured sparsity. Finally, the learned equivariant maps are validated in example recognition tasks.

\paragraph{Sparse regression.} The first experiment (Fig.~\ref{f:regression}) explores variants of the sparse regression formulation~\eqref{e:objective}. The goal is to learn a mapping $M_g=(A_g,\bb_g)$ that predicts the effect of selected image transformations $g$ on the HOG features of an image. For each transformation, the mapping $M_g$ is learned from 1,000 training images by minimising the regularised empirical risk~\eqref{e:objective2}. The performance is measured as the average Hellinger's distance $\|\phi(g\bx)-M_g\phi(\bx)\|_\text{Hell.}$ on a test set of further 1,000 images.\footnote{The Hellinger's distance $(\sum_i (\sqrt{x_i}-\sqrt{y_i})^2)^{1/2}$ is preferred to the Euclidean distance as the HOG features are histograms.} Images are randomly sampled from the ILSVRC12 train and validation datasets respectively.

This experiment focuses on predicting a small array of $5\times 5$ of HOG cells, which allows to train full regression matrices even with naive baseline regression algorithms. Furthermore, the $5\times 5$ array is predicted from a larger $9 \times 9$ input array to avoid boundary issues when images are rotated or rescaled. Both these restrictions will be relaxed later. Fig.~\ref{f:regression} compares the following methods to learn $M_g$: choosing the identity transformation $M_g=\mathbf{1}$, learning $M_g$ by optimising the objective~\eqref{e:objective} without regularisation (Least Square -- LS), with the Frobenius norm regulariser for different values of $\lambda$ (Ridge Regression -- RR), and with the sparsity-inducing regulariser~\eqref{e:ff} (Forward-Selection -- FS, using \cite{sjostrand2012spasm}) for a different number $k$ of regression coefficients per output dimension.

As can be seen in~Fig.~\ref{fig:hog_mg_rot},~\ref{fig:hog_mg_sc}, LS overfits badly, which is not surprising given that $M_g$ contains 1M parameters even for these small HOG arrays. RR performs significantly better, but it is easily outperformed by FS, confirming the very sparse nature of the solution (\eg for $k=5$ just 0.2\% of the 1M coefficients are non-zero). The best result is obtained by FS with $k=5$. As expected, the prediction error of FS is zero for a $180\degree$ rotation as this transformation is exact (Sect.~\ref{s:learning}), but note that LS and RR fail to recover it. As one might expect, errors are smaller for transformations close to identity, although in the case of FS the error remains small throughout the range.

\begin{figure*}[th!]
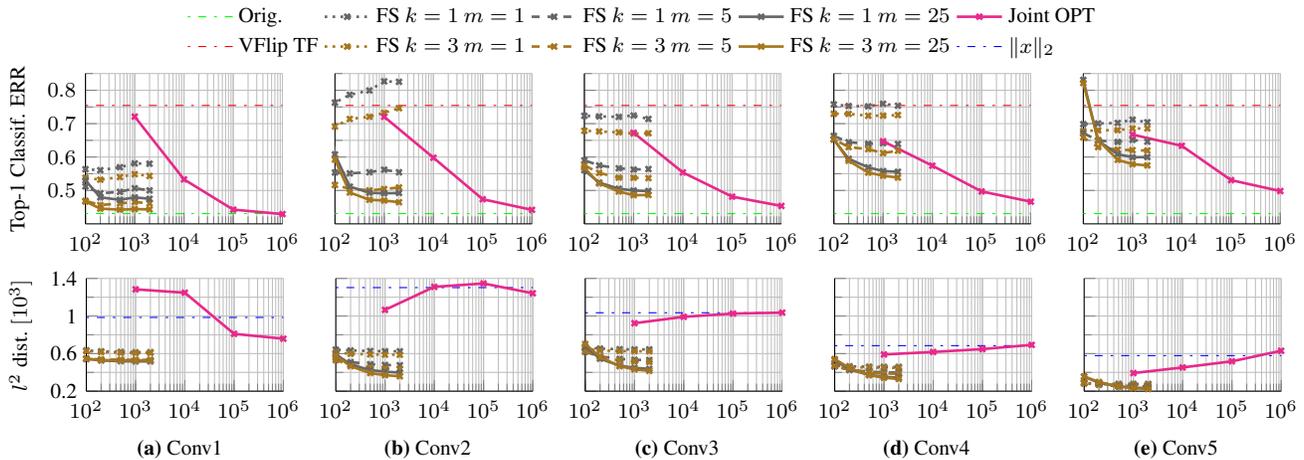

	\newlength{\cnncolw}
	\setlength{\cnncolw}{0.15\textwidth}
	\newlength{\cnncolh}
	\setlength{\cnncolh}{2cm}
	\newlength{\cnncolhs}
	\setlength{\cnncolhs}{1.5cm}
	\newlength{\cnncolhsm}
	\setlength{\cnncolhsm}{-1.2em}
	\centering
    \vspace*{-2.5em}
    \subfloat{
    	\begin{tikzpicture}
		\definecolor{mycolor1}{rgb}{0.65098,0.46275,0.11373}%
		\definecolor{mycolor2}{rgb}{0.90588,0.16078,0.54118}%
    	\begin{customlegend}[legend columns=5,
    	legend style={align=left,draw=none},
    	legend cell align=left,
    	legend entries={Orig., FS $k=1~m=1$, FS $k=1~m=5$, FS $k=1~m=25$, Joint OPT, VFlip TF, FS $k=3~m=1$, FS $k=3~m=5$, FS $k=3~m=25$, $\| x \|_2$}]
    	\addlegendimage{color=green,dash pattern=on 1pt off 3pt on 3pt off 3pt} % Orig
    	\addlegendimage{color=white!40!black,dotted,line width=1.0pt,mark size=1.5pt,mark=x,mark options={solid}} %FS gp1m1
	    \addlegendimage{color=white!40!black,dashed,line width=1.0pt,mark size=1.5pt,mark=x,mark options={solid}} % FS gp1m5 
       	\addlegendimage{color=white!40!black,solid,line width=1.0pt,mark size=1.5pt,mark=x,mark options={solid}} % FS gp1m25
    	
    	\addlegendimage{color=mycolor2,solid,line width=1.0pt,mark size=1.5pt,mark=x,mark options={solid}} % JOPT
    	
    	\addlegendimage{color=red,dash pattern=on 1pt off 3pt on 3pt off 3pt}   % TF
    	\addlegendimage{color=mycolor1,dotted,line width=1.0pt,mark size=1.5pt,mark=x,mark options={solid}} % FS gp3m1
    	\addlegendimage{color=mycolor1,dashed,line width=1.0pt,mark size=1.5pt,mark=x,mark options={solid}} % FS gp3m5
    	\addlegendimage{color=mycolor1,solid,line width=1.0pt,mark size=1.5pt,mark=x,mark options={solid}} % FS gp3m25
    	\addlegendimage{color=blue,dash pattern=on 1pt off 3pt on 3pt off 3pt} % ||x||
    	\end{customlegend}
    	\end{tikzpicture}
    	\label{fig:cnn_gprior_legend}}
    \\ \vspace*{-1.5em}
	\makebox[\textwidth][c]{
		\subfloat{
			\setlength{\figureheight}{\cnncolh}
			\setlength{\figurewidth}{\cnncolw}
			\input{figures/cnn_gprior_optres_conv1.tkiz}
			\label{fig:cnn_gprior_conv1}} \hspace*{\cnncolhsm}
		\subfloat{
			\setlength{\figureheight}{\cnncolh}
			\setlength{\figurewidth}{\cnncolw}
			\input{figures/cnn_gprior_optres_conv2.tkiz}
			\label{fig:cnn_gprior_conv2}} \hspace*{\cnncolhsm}
		\subfloat{
			\setlength{\figureheight}{\cnncolh}
			\setlength{\figurewidth}{\cnncolw}
			\input{figures/cnn_gprior_optres_conv3.tkiz}
			\label{fig:cnn_gprior_conv3}} \hspace*{\cnncolhsm}
		\subfloat{
			\setlength{\figureheight}{\cnncolh}
			\setlength{\figurewidth}{\cnncolw}
			\input{figures/cnn_gprior_optres_conv4.tkiz}
			\label{fig:cnn_gprior_conv4}} \hspace*{\cnncolhsm}
		\subfloat{
			\setlength{\figureheight}{\cnncolh}
			\setlength{\figurewidth}{\cnncolw}
			\input{figures/cnn_gprior_optres_conv5.tkiz}
			\label{fig:cnn_gprior_conv5}}
	} \\
	\vspace*{-1em}
	\setcounter{subfigure}{0}
	\makebox[\textwidth][c]{
		\captionsetup[subfigure]{oneside,margin={2.4em,0em}}
		\subfloat[Conv1]{
			\setlength{\figureheight}{\cnncolhs}
			\setlength{\figurewidth}{\cnncolw}
			\input{figures/cnn_gprior_optres_dist_conv1_l2.tkiz}
			\label{fig:cnn_gprior_dist_conv1_l2}} \hspace*{\cnncolhsm}
		\captionsetup[subfigure]{oneside,margin={0em,0em}}
		\subfloat[Conv2]{
			\setlength{\figureheight}{\cnncolhs}
			\setlength{\figurewidth}{\cnncolw}
			\input{figures/cnn_gprior_optres_dist_conv2_l2.tkiz}
			\label{fig:cnn_gprior_dist_conv2_l2}} \hspace*{\cnncolhsm}
		\subfloat[Conv3]{
			\setlength{\figureheight}{\cnncolhs}
			\setlength{\figurewidth}{\cnncolw}
			\input{figures/cnn_gprior_optres_dist_conv3_l2.tkiz}
			\label{fig:cnn_gprior_dist_conv3_l2}} \hspace*{\cnncolhsm}
		\subfloat[Conv4]{
			\setlength{\figureheight}{\cnncolhs}
			\setlength{\figurewidth}{\cnncolw}
			\input{figures/cnn_gprior_optres_dist_conv4_l2.tkiz}
			\label{fig:cnn_gprior_dist_conv4_l2}} \hspace*{\cnncolhsm}
		\subfloat[Conv5]{
			\setlength{\figureheight}{\cnncolhs}
			\setlength{\figurewidth}{\cnncolw}
			\input{figures/cnn_gprior_optres_dist_conv5_l2.tkiz}
			\label{fig:cnn_gprior_dist_conv5_l2}}
	}
	\vspace*{-0.5em}
	\caption{{\bf Comparison of regression methods for a CNN.} Regression error of an equivariant map $M_g$ learned for vertical image flips for different layers of a CNN. FS (gray and brown lines) and the task-oriented objective (purple) are evaluated against the number of training samples. Both the task loss (top) and the feature reconstruction error (bottom) are reported. In the task loss, the green dashed line is the performance of the original classifier on the original images (best possible performance) and the red dashed line the performance of this classifier on the transformed images (worst case). In the second row, the $l^2$ reconstruction error per cell is visualised together with the baseline - average $l^2$ distance of the representation to zero vector.}\label{f:cnn-regression}\label{fig:cnn_gprior}
\end{figure*}

\paragraph{Structured sparse regression.} The conclusion of the previous experiments is that sparsity is essential to achieve good generalisation. However, learning $M_g$ directly, \eg by forward-selection or by $l^1$ regularisation, can be quite expensive even if the solution is ultimately sparse. Next, we evaluate using the \emph{structured sparsity} regulariser~\eqref{e:ss}, where each output feature is predicted from a prespecified neighbourhood of input features dependent on the image transformation $g$. Fig.~\ref{fig:hog_mg_cell} repeats the experiment of Fig.~\ref{fig:hog_mg_rot} for a $45\degree$ rotation, but this time limited to neighbourhoods of $m \times m$ input HOG cells. To be able to span larger intervals of $m$, an array of $15 \times 15$ HOG cells is used. Since spatial sparsity is now imposed {\em a-priori}, LS, RR, and FS perform nearly equivalently for $m\leq3$, with the best result achieved by FS with $k=5$ and a small neighbourhood of $m = 3$ cells. There is also a significant computational advantage in structured sparsity (Tab.~\ref{tab:gprior-time}) as it limits the effective size of the regression problems to be solved. We conclude that structured sparsity is highly preferable over generic sparsity.

\paragraph{Regression quality.} So far results have been given in term of the reconstruction error of the features; this paragraph relates this measure to the practical performance of the learned mappings. The first experiment is qualitative and uses the HOGgle technique~\cite{vondrick13hoggles:} to visualise the transformed features. As shown in Fig.~\ref{fig:hoggles}, the visualisations of $\phi(g\bx)$ and $M_g \phi(\bx)$ are indeed nearly identical, validating the mapping $M_g$. The second experiment (Fig.~\ref{fig:gprior_vs_dist}) evaluates instead the performance of transformed HOG features quantitatively, in a classification problem. To this end, an SVM classifier $\langle \bw, \phi(\bx) \rangle$ is trained to discriminate between dog and cat faces using the data of \cite{parkhi11truth} (using $15\times 15$ HOG templates, 400 training and 1,000 testing images evenly split among cats and dogs). Then a progressively larger rotation or scaling $g^{-1}$ is applied to the input image and the effect compensated by $M_g$, computing the SVM score as $\langle \bw, M_g \phi(g^{-1}\bx) \rangle$ (equivalently the model is transformed by $M_g^\top$). The performance of the compensated classifier is nearly identical to the original classifier for all angles and scales, whereas the uncompensated classifier $\langle \bw, \phi(g^{-1}\bx)\rangle$ rapidly fails, particularly for rotation. We conclude that equivariant transformations encode visual information effectively.

% ----------------------------------------------------------
\subsection{Equivariance in deep representations}\label{s:experiments-deep}
% ----------------------------------------------------------

\begin{figure}
	\begin{center}
		%		\fbox{\rule{0pt}{2in} \rule{.9\linewidth}{0pt}}
		\setlength{\figureheight}{3cm}
		\setlength{\figurewidth}{0.6\linewidth}
		\input{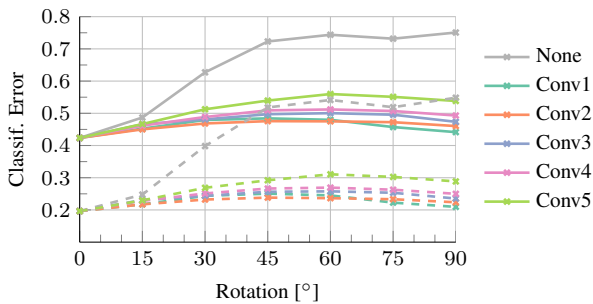}
	\end{center}
	\vspace{-2em}
\caption{{\bf Learning equivariant CNN mappings for image rotations.} The setting is similar to Fig.~\ref{f:cnn-regression}, extended to several rotations $g$ but limted to the task-oriented regression method. The solid and dashed lines report respectively the top1 and top5 errors on the ILSVRC12 validation set.}\label{fig:cnn_jointopt_rot}
\end{figure}

The previous section validated learning equivariant transformations in shallow representations such as HOG. This section extends these results to deep representations, using the \textsc{Alexn} CNN~\cite{krizhevsky12imagenet} as a reference state-of-the-art deep feature extractor using the MatConvNet framework~\cite{vedaldi2014matconvnet}. \textsc{Alexn} is the composition of twenty functions, grouped into five convolutional layers (comprising filtering, max-pooling, normalisation and ReLU) and three fully-connected layers (filtering and ReLU). The experiments look at the convolutional layers Conv1 to Conv5 right after the linear filters (learning the linear transformation layers after the ReLU was found to be harder due to the non-negativity of the features).

\paragraph{Regression methods.} The first experiment (Fig.~\ref{f:cnn-regression}) compares different methods to learn equivariant mappings $M_g$ in a CNN. The first method is FS, computed for different neighbourhood sizes $m$ and sparsity $k$. The second is the task oriented formulation of Sect.~\ref{s:regression} using a transformation layer. Both the $l^2$ reconstruction error of the features and the classification error (task-oriented loss) are reported. As in Sect.~\ref{s:transformation}, the latter is the classification error of the compensated network $\phi_2 \circ M_g \circ \phi_1(g^{-1}\bx)$ in the ImageNet ILSVCR data (the reported error is measured on the validation data, but optimised on the training data). The figure reports the evolution of the loss as more training samples are used. For the purpose of this experiment, $g$ is set to vertical image flip. Fig.~\ref{fig:cnn_jointopt_rot} repeats the experiments for the task-oriented objective and rotations $g$ from 0 to 90 degrees (the fact that intermediate rotations are slightly harder to reconstruct suggests that a better $M_g$ could be learned by addressing more carefully interpolation and boundary effects).

Several observations can be made. First, all methods perform substantially better than doing nothing ($\sim 75\%$ top-1 error), recovering most if not all the performance of the original classifier ($43\%$). This demonstrates that linear equivariant mappings $M_g$ can be learned successfully for CNNs too. Second, for the shallower features up to Conv2, FS is better: it requires less training samples and it has a smaller reconstruction error and comparable classification error than the task-oriented loss. Compared to Sect.~\ref{s:experiments-shallow}, however, the best setting $m=3$, $k=25$ is substantially less sparse. However, from Conv3 onwards, the task-oriented loss is better, converging to a much lower classification error than FS. FS still achieves a significantly smaller reconstruction error, showing that feature reconstruction is not always predictive of classification performance. Third, the classification error increases somewhat with depth, matching the intuition that deeper layers contain more specialised information: as such, perfectly transforming these layers for transformations not experienced during training (such as vertical flips) may not be possible.

%\vspace{-0.5em}
\paragraph{Testing transformations.} Next, we investigate which geometric transformations can be represented by different layers of a CNN (Tab.~\ref{tab:cnn_tfperf}), considering in particular horizontal and vertical flips, rescaling by half, and rotation of $90\degree$. First, for transformations such as horizontal flips and scaling, learning equivariant mappings is not better than leaving the features unchanged: the reason is that the CNN is implicitly learned to be invariant to such factors. For vertical flips and rotations, however, the learned equivariant mapping substantially reduce the error. In particular, the first few layers are easily transformable, confirming their generic nature.

%\vspace{-0.5em}
\paragraph{Quantifying invariance.} One use of the mapping $M_g$ is the identification of invariant features in the representation. These are the ones that are best predicted by themselves after a transformation. In practice,  a transformation layer in a CNN (Sect.~\ref{s:transformation}) identifies invariant feature \emph{channels} since the same transformation filters are applied uniformly at all spatial locations.  In practice, invariance is almost never achieved exactly; instead, the degree of invariance of a feature channel is scored as the ratio of the Euclidean norm of the corresponding row of $M_g$ with the same after suppressing the ``diagonal'' component of that row. Then, the $p$ rows of $M_g$ with the highest invariance score are replaced by (scaled) rows of the identity matrix. Finally, the performance of the modified transformation $\bar M_g$ is evaluated and accepted if the classification performance does not deteriorate by more than $5\%$ relative to $M_g$. The corresponding feature channels for the largest possible $p$ are then be considered approximately invariant.

Table~\ref{tab:cnn-invariance} reports the result of this analysis for horizontal and vertical flips, rescaling, and $90\degree$ rotation in the \textsc{Alexn} CNN. There are several notable observations. First, for transformations for which the network is overall invariant such as horizontal flips and rescaling, invariance is obtained largely in Conv3 or Conv4. Second, invariance is not always increasing with depth, as for example Conv1 tends to be more invariant than Conv2. This is possible because, even if the feature channels in a layer are invariant, the spatial pooling in the subsequent layer may not be. Third, the number of invariant features is significantly smaller for unexpected transformations such as vertical flips and $90\degree$ rotations, further validating the approach.

\begin{table}
	\tabcolsep=0.11cm
	\footnotesize
	\begin{center}

\begin{tabular}{| l |  c c |  c c |  c c |  c c | }\hline
	\multirow{2}{*}{Layer} & \multicolumn{2}{c |}{Horiz. Flip} & \multicolumn{2}{c |}{Vert. Flip} & \multicolumn{2}{c |}{Sc. $2^{-\frac{1}{2}}$} & \multicolumn{2}{c |}{Rot. $90\degree$}  \\
	& Top1 & Top5  & Top1 & Top5  & Top1 & Top5  & Top1 & Top5  \\ \hline
	\textbf{None}  & 0.44 & 0.21  & 0.75 & 0.54  & 0.61 & 0.37  & 0.75 & 0.54  \\ \hline
	\textbf{Conv1}  & 0.43 & 0.20  & 0.43 & 0.20  & 0.45 & 0.22  & 0.44 & 0.20  \\
	\textbf{Conv2}  & 0.45 & 0.22  & 0.46 & 0.22  & 0.48 & 0.24  & 0.46 & 0.22  \\
	\textbf{Conv3}  & 0.45 & 0.21  & 0.46 & 0.22  & 0.49 & 0.25  & 0.47 & 0.23  \\
	\textbf{Conv4}  & 0.44 & 0.21  & 0.48 & 0.24  & 0.49 & 0.25  & 0.49 & 0.25  \\
	\textbf{Conv5}  & 0.44 & 0.21  & 0.51 & 0.26  & 0.50 & 0.26  & 0.53 & 0.28  \\
	\hline
\end{tabular}

\end{center}
	\vspace{-2em}
	\caption{{\bf CNN equivariance.} Performance on the ILSVRC12 validation set of compensated CNN classifier using learned equivariant mappings for selected transformations. For reference, the top-1 and top-5 error of the unmodified \textsc{Alexn} are $0.43$ and $0.20$ respectively.}\label{tab:cnn_tfperf}
	\tabcolsep=0.11cm
	\footnotesize
	\vspace{-1em}
	\begin{center}

\begin{tabular}{| l |  r r |  r r |  r r |  r r | }\hline
	\multirow{2}{*}{Layer} & \multicolumn{2}{c |}{Horiz. Flip} & \multicolumn{2}{c |}{Vert. Flip} & \multicolumn{2}{c |}{Sc. $2^{-\frac{1}{2}}$} & \multicolumn{2}{c |}{Rot. $90\degree$}  \\ 
	& Num & \%  & Num & \%  & Num & \%  & Num & \%  \\ \hline 
	\textbf{Conv1}  & 52 & 54.17  & 53 & 55.21  & 95 & 98.96  & 42 & 43.75  \\ 
	\textbf{Conv2}  & 131 & 51.17  & 45 & 17.58  & 69 & 26.95  & 27 & 10.55  \\ 
	\textbf{Conv3}  & 238 & 61.98  & 132 & 34.38  & 295 & 76.82  & 120 & 31.25  \\ 
	\textbf{Conv4}  & 343 & 89.32  & 124 & 32.29  & 378 & 98.44  & 101 & 26.30  \\ 
	\textbf{Conv5}  & 255 & 99.61  & 47 & 18.36  & 252 & 98.44  & 56 & 21.88  \\ 
	\hline 
\end{tabular}

\end{center}
	\vspace{-2em}
	\caption{{\bf CNN invariance.} Number and percentage of invariant feature channels in the \textsc{Alexn} network, identified by analysing corresponding equivariant transformations.}
	\label{tab:cnn-invariance}
\vspace{-1em}
\tabcolsep=0.11cm
\footnotesize
\begin{center}

\begin{tabular}{| l |  r r |  r r |  r r | }\hline
	\multirow{2}{*}{Layer} & \multicolumn{2}{c|}{\textsc{Imnet} $\to$ \textsc{Alexn}} & \multicolumn{2}{c|}{\textsc{Plcs} $\to$ \textsc{Alexn}} & \multicolumn{2}{c|}{\textsc{Plcs-h} $\to$ \textsc{Alexn}}  \\ 
	& Top1 & Top5  & Top1 & Top5  & Top1 & Top5  \\ \hline 
% For conv1-imnet->alexnet used results of epoch 4 (started to overfit, fell down by a percent to 0.44 & 0.21 -> early stopping), the rest epoch 6
	\textbf{Conv1}  & 0.43 & 0.20  & 0.43 & 0.20  & 0.43 & 0.20  \\ 
	\textbf{Conv2}  & 0.46 & 0.22  & 0.47 & 0.23  & 0.46 & 0.22  \\ 
	\textbf{Conv3}  & 0.46 & 0.22  & 0.50 & 0.25  & 0.47 & 0.23  \\ 
	\textbf{Conv4}  & 0.46 & 0.22  & 0.54 & 0.29  & 0.49 & 0.24  \\ 
	\textbf{Conv5}  & 0.50 & 0.25  & 0.65 & 0.39  & 0.52 & 0.27  \\ 
	\hline 
\end{tabular}

\end{center}
\vspace{-2em}
\caption{{\bf CNN equivalence.} Performance on the ILSVRC12 validation set of several ``Franken-CNNs'' obtained by stitching the first portion of \textsc{Imnet}, \textsc{Plcs} and \textsc{Plcs-h} up to a certain convolutional layer and the last portion of \textsc{Alexn}.}\label{tab:frankennet_layers}
\end{table}

%\begin{figure}
%	\begin{center}
%		%		\fbox{\rule{0pt}{2in} \rule{.9\linewidth}{0pt}}
%		\setlength{\figureheight}{2.5cm}
%		\setlength{\figurewidth}{0.33\linewidth}
%		\input{figures/cnn_gprior_optres.tkiz}
%	\end{center}
%	\vspace{-2em}
%	\caption{Error rates of different regression methods used to learn equivariant mapping of the \textsc{Imnet} CNN for vertical image flips.}\label{fig:cnn_gprior}
%\end{figure}
%
%
%\begin{figure}
%	\begin{center}
%		%		\fbox{\rule{0pt}{2in} \rule{.9\linewidth}{0pt}}
%		\setlength{\figureheight}{2.5cm}
%		\setlength{\figurewidth}{0.33\linewidth}
%		\input{figures/cnn_gprior_optres_dist.tkiz}
%	\end{center}
%	\vspace{-2em}
%	\caption{Comparison of the same regression methods on the same task as in \ref{fig:cnn_gprior}, however measured with average Hellinger distance per a spatial cell.}\label{fig:cnn_gprior_dist}
%\end{figure}

% ----------------------------------------------------------
\subsection{Equivalence of deep representations}\label{s:experiments-equivalence}
% ----------------------------------------------------------

While the previous two sections studied the equivariance of representations, this section looks at their equivalence. The goal is to clarify whether heterogeneous representations may in fact capture the same visual information by replacing part of a representation with another using the methods of Sect.~\ref{s:learning} and Sect.~\ref{s:stitching}.

To validate this idea, the first several layers $\phi_1'$ of the \textsc{Alexn} CNN $\phi'=\phi_2'\circ \phi_1'$ are swapped with layers $\phi_1$ from \textsc{Imnet}, also trained on the ILSVRC12 data, \textsc{Plcs}~\cite{zhou2014places}, trained on the MIT Places data, and \textsc{Plcs-h}, trained on a mixture of MIT Places and ILSVRC12 images. These representations have a similar, but not identical, structure and entirely different parametrisations.

Table~\ref{tab:frankennet_layers} shows the top-1 performance of hybrid models $\phi_2' \circ E_{\phi_1\rightarrow\phi_1'} \circ \phi_1$, where the equivalence map $E_{\phi_1\rightarrow\phi_1'}$ is learned as a stitching layer (Sect.~\ref{s:stitching}) from ILSVRC12 training images. There are a number of notable facts. First, setting $E_{\phi\rightarrow\phi'}=\mathbf{1}$ to the identity map has a top-1 error $>99\%$ (not shown in the table), matching the intuition that different parametrisations make feature channels not directly compatible. Second, a very good level of equivalence can be established up to Conv4 between \textsc{Alexn} and \textsc{Imnet}, and a slightly less good one between \textsc{Alexn} and \textsc{Plcs-h}; however, in \textsc{Plcs} deeper layers are substantially less compatible. Specifically, Conv1 and Conv2 are interchangeable in all cases, whereas Conv5 is not fully interchangeable, particularly for \textsc{Plcs}. This corroborates the intuition that Conv1 and Conv2 are generic image codes, whereas Conv5 is more task-specific. Note however that, even in the worst case, performance is dramatically better than chance, demonstrating that all such features are compatible to an extent.

\subsection{Application to structured-output regression}\label{s:so}

As a complement of the theoretical investigation so far, this section shows a direct practical application of the learned equivariant mappings of Sect.~\ref{s:learning} to structured-output regression~\cite{taskar03max-margin}. In structured regression an input image $\bx$ is mapped to a label $\by$ by the function $\hat\by(\bx) = \operatorname{argmax}_{\by,\bz} \langle \phi(\bx,\by,\bz), \bw \rangle$ (direct regression) where $\bz$ is an optional latent variable and $\phi$ a joint feature map. If $\by$ and/or $\bz$ include geometric parameters, the joint feature can be partially of fully rewritten as $\phi(\bx,\by,\bz)= M_{\by,\bz} \phi(\bx)$, reducing inference to the maximisation of $\langle M_{\by,\bz}^\top \bw, \phi(\bx)\rangle$ (equivariant regression). There are two computational advantages: (i) the representation $\phi(\bx)$ needs to be computed just once and (ii) the vectors $M_{\by,\bz}^\top \bw$ can be precomputed.

\begin{figure}[h!]
	\vspace{-1em}
	\begin{center}
		\begin{footnotesize}
			\setlength{\tabcolsep}{1pt}
			\newcommand{\CatInsIm}[2]{\includegraphics[width=0.19\linewidth,clip=true,#2]{figures/cats/#1}}
			\newcommand{\CatRotImN}[2]{\CatInsIm{sel_nice/rot_cath_000#1.pdf}{#2}}
			\newcommand{\CatPosImN}[2]{\CatInsIm{sel_nice/pos_cath_000#1.pdf}{#2}}
			\newcommand{\CatRotImU}[2]{\CatInsIm{sel_ugly/rot_cath_000#1.pdf}{#2}}
			\newcommand{\CatPosImU}[2]{\CatInsIm{sel_ugly/pos_cath_000#1.pdf}{#2}}
			\begin{tabular}{ c c c c c}
				\CatRotImN{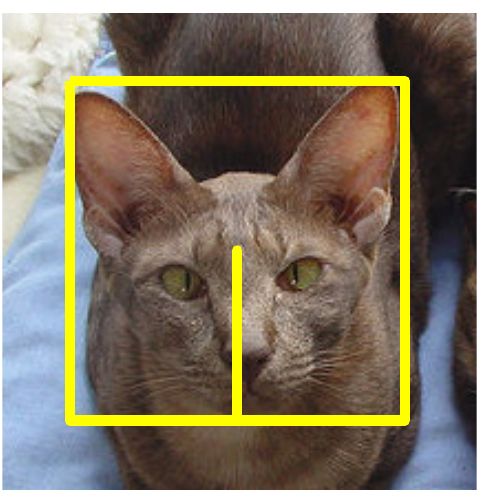}{} & \CatRotImN{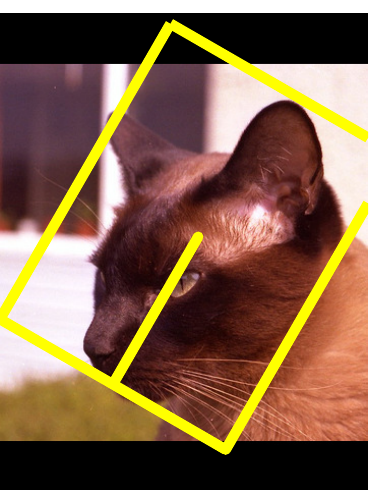}{trim=0 2em 0 2.5em}    & \CatRotImN{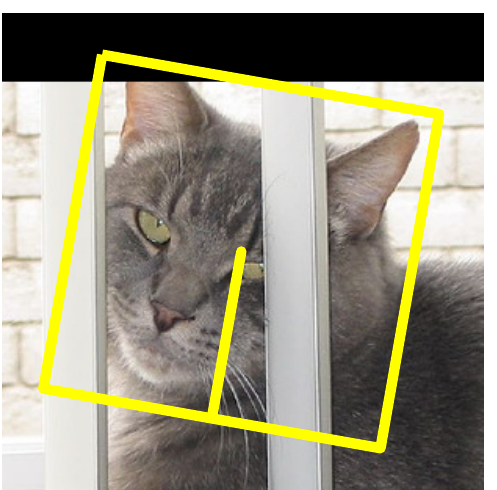}{trim=1em 0 1em 3.1em} & \CatRotImN{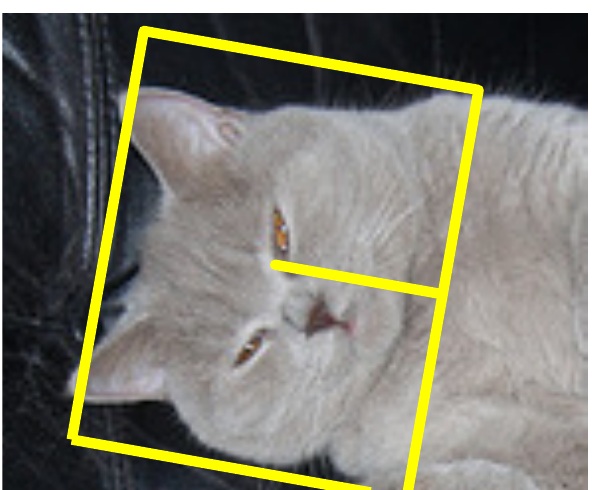}{}   & \CatRotImU{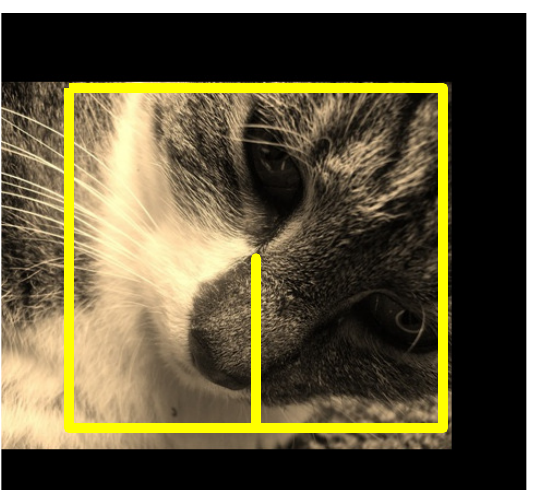}{trim=0 2em 2.7em 2.9em}    \\
				\CatPosImN{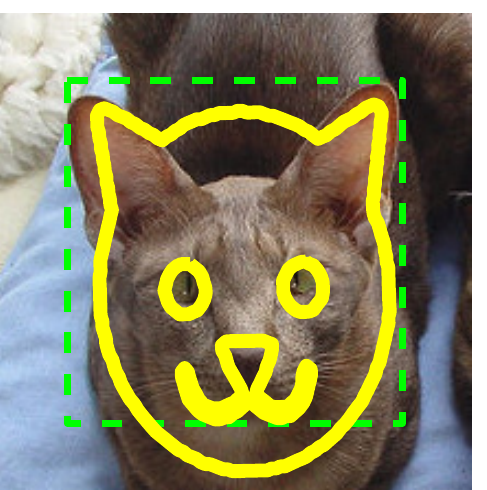}{} & \CatPosImN{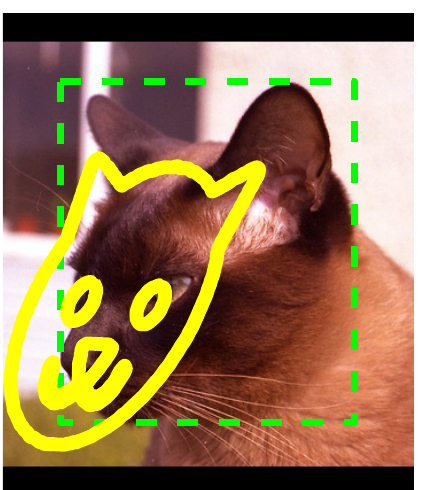}{trim=0 1em 0 1.6em}    & \CatPosImN{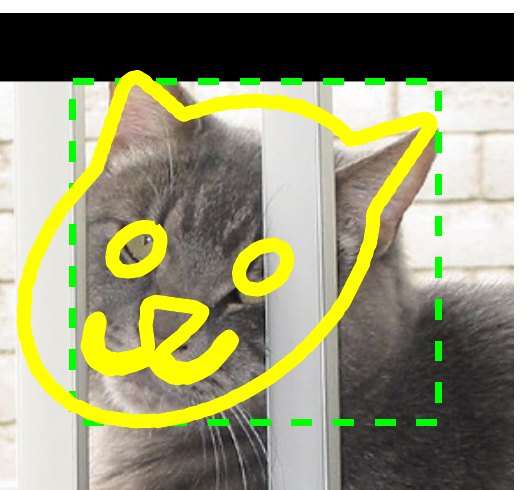}{trim=1em 0 1em 3.1em} & \CatPosImN{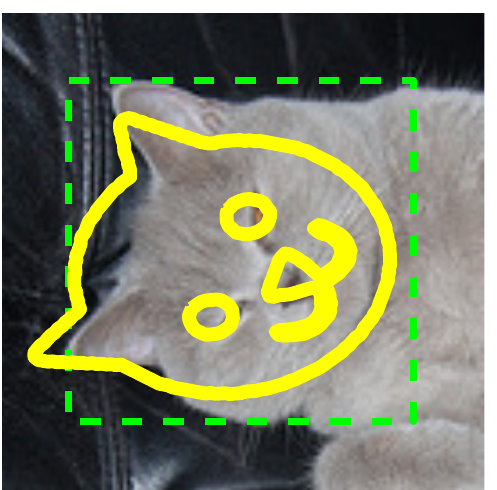}{}   & \CatPosImU{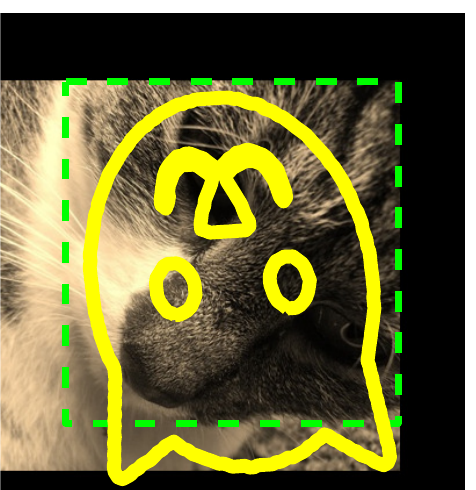}{trim=0 2em 2.7em 2.9em} \\

			\end{tabular}
			\vspace{-1em}
			\caption{{\bf Equivariant regression examples.} Rotation (top) and affine pose (bottom) prediction for cat faces in the VOC07 parts data. The estimated affine pose is represented by eyes and nose location. The first four columns contain examples of successful regressions an the last a failure case. Regression uses the CNN Conv5 features computed within the green dashed box.}
			\label{fig:cattf_examples}
		\end{footnotesize}
	\end{center}
%	\vspace{-2em}
\end{figure}

This idea is demonstrated on the task of pose estimation, where $\by = g$ is a geometric transformation in a class $g^{-1}\in G$ of possible poses of an object. As an example, consider estimating the pose of cat faces in the PASCAL VOC 2007~(VOC07)~\cite{everingham07pascal} data using for $G$ either (i) rotations or (ii) affine transformations (Fig.~\ref{fig:cattf_examples}). The rotations in $G$ are sampled uniformly every 10 degrees and the ground-truth rotation of a face is defined by the line connecting the nose to the midpoints between the eyes. These keypoints are obtained as the center of gravity of the corresponding regions in the VOC07 part annotations~\cite{chen_cvpr14}. The affine transformations in $G$ are obtained instead by clustering the vectors $[\mathbf{c}_l^\top, \mathbf{c}_r^\top,\mathbf{c}_n^\top]^\top$ containing the location of eyes and nose of 300 example faces in the VOC07 data. 
The clusters are obtained using GMM-EM on the training data and used to map the test data to the same pose classes for evaluation. $G$ then contains the set of affine transformations mapping the keypoints $[\bar{\mathbf{c}}_l^\top, \bar{\mathbf{c}}_r^\top, \bar{\mathbf{c}}_n^\top]^\top$ in a canonical frame to each cluster center.

The matrices $M_g$ are pre-learned (from generic images, not containing cats) using FS with $k=5$ and $m=3$ as in Sect.~\ref{s:learning}. Since cat faces in VOC07 data are usually upright, a second more challenging version of the data (denoted by the symbol $\circlearrowleft$) augmented with random image rotations is considered as well. The direct $\langle  \bw, \phi(g\bx)\rangle$ and equivariant $\langle  \bw, M_{g}\phi(\bx)\rangle$ scoring functions are learned using 300 training samples and evaluated on 300 test ones.

\begin{figure}[h!]
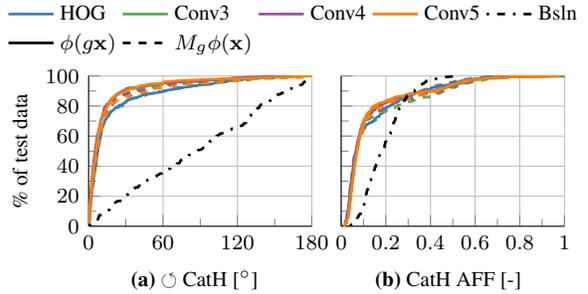

\vspace{-1em}
	\newlength{\catrotcolw}
	\setlength{\catrotcolw}{0.17\textwidth}
	\newlength{\catrotcolh}
	\setlength{\catrotcolh}{2cm}
	\newlength{\catrotcolhsm}
	\setlength{\catrotcolhsm}{-1.8em}
%	\vspace*{-1em}
	%\hspace*{-1em}
	\begin{center}
	%\vspace*{-1.2em}
	%\hspace*{-3em}
	
	\subfloat{
	\footnotesize
   	\begin{tikzpicture}
	\definecolor{mycolor1}{rgb}{0.21569,0.49412,0.72157}%
	\definecolor{mycolor2}{rgb}{0.30196,0.68627,0.29020}%
	\definecolor{mycolor3}{rgb}{0.59608,0.30588,0.63922}%
	\definecolor{mycolor4}{rgb}{1.00000,0.49804,0.00000}%
   	\begin{customlegend}[legend columns=5,
   	legend style={align=left,draw=none},
   	legend cell align=left,
   	legend entries={HOG, Conv3, Conv4, Conv5, Bsln, $\phi(g\mathbf{x})$, $M_g \phi(\mathbf{x})$}]
   	\addlegendimage{color=mycolor1,solid,line width=1.0pt} % HOG
   	\addlegendimage{color=mycolor2,solid,line width=1.0pt} % Conv3
   	\addlegendimage{color=mycolor3,solid,line width=1.0pt} % Conv4 
   	\addlegendimage{color=mycolor4,solid,line width=1.0pt} % Conv5
   	
   	\addlegendimage{color=black,dash pattern=on 1pt off 3pt on 3pt off 3pt,line width=1.0pt} % Bsln
   	\addlegendimage{color=black,solid,line width=1.0pt}
   	\addlegendimage{color=black,dashed,line width=1.0pt}
   	\end{customlegend}
   	\end{tikzpicture}
   	\label{fig:cattf_legend}}
    \vspace*{-1.2em}
	\hspace*{-1.2em}
	\setcounter{subfigure}{0}
	%\makebox[\textwidth][c]{
		%\subfloat[CatH]{
		%	\setlength{\figureheight}{\catrotcolh}
		%	\setlength{\figurewidth}{\catrotcolw}
		%	\input{figures/catrot_errc_cat.tkiz}
		%	\label{fig:catrot_errc_cat-head}} \hspace*{\catrotcolhsm}
		\captionsetup[subfigure]{oneside,margin={2.3em,0em}}
		\subfloat[{$\circlearrowleft$ CatH [$\degree$]}]{
			\setlength{\figureheight}{\catrotcolh}
			\setlength{\figurewidth}{\catrotcolw}
			\input{figures/catrot_errc_cat_rrot.tkiz}
			\label{fig:catrot_errc_cat_rrot}} \hspace*{\catrotcolhsm}
		\captionsetup[subfigure]{oneside,margin={0.3em,0em}}
		\subfloat[{CatH AFF [-]}]{
			\setlength{\figureheight}{\catrotcolh}
			\setlength{\figurewidth}{\catrotcolw}
			\input{figures/catpos_errc_cat-head.tkiz}
			\label{fig:catpos_errc_cat-head}} 
	%}
	\end{center}
	\vspace{-2em}
	\caption{{\bf Equivariant regression errors.} Cumulative error curves for the rotation and affine pose regressors of Table~\ref{tab:hog_catrot}.}\label{fig:cattf_errc}
\end{figure}

Table~\ref{tab:hog_catrot} reports the accuracy and speed obtained for HOG and CNN Conv3, Conv4, and Conv5 features for direct and equivariant regression. The latter is generally as good or nearly as good as direct regression, but up to 22 times faster validating once more the mappings $M_g$. Fig.~\ref{fig:cattf_errc} shows the cumulative error curves for the different regressors.

\begin{table}[th!]
	\tabcolsep=0.11cm
	\footnotesize
\begin{center}
\begin{tabular}{| l | c |  c c |  c c |  c c |  c c | }\hline
	\multirow{2}{*}{$\phi(x)$} &  & \multicolumn{2}{c |}{ HOG } & \multicolumn{2}{c |}{ Conv3 } & \multicolumn{2}{c |}{ Conv4 } & \multicolumn{2}{c |}{ Conv5 }  \\ 
	& Bsln  & $g$ & $M_{g}$  & $g$ & $M_{g}$  & $g$ & $M_{g}$  & $g$ & $M_{g}$  \\ \hline 
	\textbf{Rot} [$\degree$] & 23.8  & 14.9 & 17.0   & 13.3 & 11.6   & 10.5 & 11.1   & 10.1 & 13.4   \\ 
	\textbf{Rot $\circlearrowleft$} [$\degree$] & 86.9  & 18.9 & 19.1   & 13.2 & 15.0   & 12.8 & 15.3   & 12.9 & 17.4   \\ 
	\hline 
	\textbf{Aff} [-] & 0.35  & 0.25 & 0.25  & 0.25 & 0.28  & 0.24 & 0.26  & 0.24 & 0.26  \\ 
	\hline
	\textbf{Time/TF} [ms] & -  & 18.2 & 0.8  & 59.4 & 6.9  & 65.0 & 7.0  & 70.1 & 5.7  \\ 
	\textbf{Speedup} [-] & -  & 1 & 21.9  & 1 & 8.6  & 1 & 9.3  & 1 & 12.3  \\ 
	\hline \end{tabular}
\end{center}
	\vspace{-2em}
	\caption{{\bf Equivariant regression.} The table reports the prediction errors for the cat head rotation/affine pose with direct/equivariant structured SVM regressors. The error is measured in expected degrees of residual rotation or as the average keypoint distance in the normalised face frame, respectively. The baseline method predicts a constant transformation.}
	\label{tab:hog_catrot}
\end{table}

%\vspace{-0.5em}
% ----------------------------------------------------------
\section{Summary}\label{s:summary}
% ----------------------------------------------------------
%\vspace{-0.5em}
This paper introduced the idea of studying representations by learning their equivariant and equivalence properties. It was shown that shallow representations and the first several layers of deep state-of-the-art CNNs transform in an easily predictable manner with image warps and that they are interchangeable, and hence equivalent, in different architectures. Deeper layers share some of these properties but to a lesser degree, being more task-specific.  In addition to the use as analytical tools, these methods have practical applications such as accelerating structured-output regressors classifier in a simple and elegant manner.

\paragraph{Acknowledgments.} Karel Lenc was partially supported by an Oxford Engineering Science DTA.
%This work only begins this new line of research. It is likely that the underlying regression technology can be improved, for example by looking at interpolating features more carefully in permutation layers or by learning non-linear regressor. Furthermore, finding a feature reconstruction metric that directly correlates with classification performance remains an open problem.

% --------------------------------------------------------------------
\vspace{-0.em}
\footnotesize
\bibliographystyle{ieee}
\bibliography{bibliography/vedaldi,literature}

\begin{thebibliography}{10}\itemsep=-1pt

\bibitem{chatfield14return}
K.~Chatfield, K.~Simonyan, A.~Vedaldi, and A.~Zisserman.
\newblock Return of the devil in the details: Delving deep into convolutional
  nets.
\newblock In {\em Proc. {BMVC}}, 2014.

\bibitem{chen_cvpr14}
X.~Chen, R.~Mottaghi, X.~Liu, S.~Fidler, R.~Urtasun, and A.~Yuille.
\newblock Detect what you can: Detecting and representing objects using
  holistic models and body parts.
\newblock In {\em IEEE Conference on Computer Vision and Pattern Recognition
  (CVPR)}, 2014.

\bibitem{csurka04visual}
G.~Csurka, C.~R. Dance, L.~Dan, J.~Willamowski, and C.~Bray.
\newblock Visual categorization with bags of keypoints.
\newblock In {\em Proc. {ECCV} Workshop on Stat. Learn. in Comp. Vision}, 2004.

\bibitem{dalal05histograms}
N.~Dalal and B.~Triggs.
\newblock Histograms of oriented gradients for human detection.
\newblock In {\em Proc. {CVPR}}, 2005.

\bibitem{denil13predicting}
M.~Denil, , B.~Shakibi, L.~Dinh, M.~Ranzato, and N.~de~Freitas.
\newblock Predicting parameters in deep learning.
\newblock In {\em Proc. {NIPS}}, 2013.

\bibitem{Donahue2013}
J.~Donahue, Y.~Jia, O.~Vinyals, J.~Hoffman, N.~Zhang, E.~Tzeng, and T.~Darrell.
\newblock Decaf: {A} deep convolutional activation feature for generic visual
  recognition.
\newblock {\em CoRR}, abs/1310.1531, 2013.

\bibitem{everingham07pascal}
M.~Everingham, A.~Zisserman, C.~Williams, and L.~V. Gool.
\newblock The {PASCAL} visual obiect classes challenge 2007 ({VOC2007})
  results.
\newblock Technical report, Pascal Challenge, 2007.

\bibitem{goodfellow09}
I.~Goodfellow, H.~Lee, Q.~V. Le, A.~Saxe, and A.~Y. Ng.
\newblock Measuring invariances in deep networks.
\newblock In {\em Advances in neural information processing systems}, pages
  646--654, 2009.

\bibitem{jegou10aggregating}
H.~J{\'e}gou, M.~Douze, C.~Schmid, and P.~P{\'e}rez.
\newblock Aggregating local descriptors into a compact image representation.
\newblock In {\em Proc. {CVPR}}, 2010.

\bibitem{krizhevsky12imagenet}
A.~Krizhevsky, I.~Sutskever, and G.~E. Hinton.
\newblock Imagenet classification with deep convolutional neural networks.
\newblock In {\em Proc. {NIPS}}, 2012.

\bibitem{leung01representing}
T.~Leung and J.~Malik.
\newblock Representing and recognizing the visual appearance of materials using
  three-dimensional textons.
\newblock {\em {IJCV}}, 43(1), 2001.

\bibitem{lindeberg98principles}
T.~Lindeberg.
\newblock Principles for automatic scale selection.
\newblock Technical Report ISRN KTH/NA/P 98/14 SE, Royal Institute of
  Technology, 1998.

\bibitem{lowe99object}
D.~G. Lowe.
\newblock Object recognition from local scale-invariant features.
\newblock In {\em Proc. {ICCV}}, 1999.

\bibitem{lowe04distinctive}
D.~G. Lowe.
\newblock Distinctive image features from scale-invariant keypoints.
\newblock {\em {IJCV}}, 2(60):91--110, 2004.

\bibitem{mikolajc03survey}
K.~Mikolajczyk and C.~Schmid.
\newblock A performance evaluation of local descriptors.
\newblock In {\em Proc. {CVPR}}, 2003.

\bibitem{parkhi11truth}
O.~Parkhi, A.~Vedaldi, C.~V. Jawahar, and A.~Zisserman.
\newblock The truth about cats and dogs.
\newblock In {\em Proc. {ICCV}}, 2011.

\bibitem{perronnin06fisher}
F.~Perronnin and C.~Dance.
\newblock {F}isher kernels on visual vocabularies for image categorizaton.
\newblock In {\em Proc. {CVPR}}, 2006.

\bibitem{Razavian2014}
A.~S. Razavian, H.~Azizpour, J.~Sullivan, and S.~Carlsson.
\newblock {CNN} features off-the-shelf: an astounding baseline for recognition.
\newblock In {\em {CVPR} DeepVision Workshop}, 2014.

\bibitem{ILSVRCarxiv14}
O.~Russakovsky, J.~Deng, H.~Su, J.~Krause, S.~Satheesh, S.~Ma, Z.~Huang,
  A.~Karpathy, A.~Khosla, M.~Bernstein, A.~C. Berg, and L.~Fei-Fei.
\newblock Imagenet large scale visual recognition challenge, 2014.

\bibitem{schimdt12learning}
U.~Schimdt and S.~Roth.
\newblock Learning rotation-aware features: From invariant priors to
  equivariant descriptors.
\newblock In {\em Proc. {CVPR}}, 2012.

\bibitem{sermanet14overfeat:}
P.~Sermanet, D.~Eigen, X.~Zhang, M.~Mathieu, R.~Fergus, and Y.~LeCun.
\newblock Overfeat: Integrated recognition, localization and detection using
  convolutional networks.
\newblock volume abs/1312.6229, 2014.

\bibitem{sifre13rotation}
L.~Sifre and S.~Mallat.
\newblock Rotation, scaling and deformation invariant scattering for texture
  discrimination.
\newblock In {\em Proc. {CVPR}}, 2013.

\bibitem{simonyan2013deep}
K.~Simonyan, A.~Vedaldi, and A.~Zisserman.
\newblock Deep inside convolutional networks: Visualising image classification
  models and saliency maps.
\newblock In {\em ICLR Workshop}, 2013.

\bibitem{sivic03video}
J.~Sivic and A.~Zisserman.
\newblock Video {Google}: A text retrieval approach to object matching in
  videos.
\newblock In {\em Proc. {ICCV}}, 2003.

\bibitem{sjostrand2012spasm}
K.~Sj{\"o}strand, L.~H. Clemmensen, R.~Larsen, and B.~Ersb{\o}ll.
\newblock Spasm: A matlab toolbox for sparse statistical modeling.
\newblock {\em Journal of Statistical Software}, 2012.

\bibitem{sohn2012learning}
K.~Sohn and H.~Lee.
\newblock Learning invariant representations with local transformations.
\newblock {\em CoRR}, abs/1206.6418, 2012.

\bibitem{taskar03max-margin}
B.~Taskar, C.~Guestrin, and D.~Koller.
\newblock Max-margin markov networks.
\newblock In {\em Proc. {NIPS}}, 2003.

\bibitem{vedaldi10vlfeat}
A.~Vedaldi and B.~Fulkerson.
\newblock {VLFeat} -- {An} open and portable library of computer vision
  algorithms.
\newblock In {\em Proc. {ACM} Int. Conf. on Multimedia}, 2010.

\bibitem{vedaldi2014matconvnet}
A.~Vedaldi and K.~Lenc.
\newblock {MatConvNet} - convolutional neural networks for {MATLAB}.
\newblock {\em CoRR}, abs/1412.4564, 2014.

\bibitem{vondrick13hoggles:}
C.~Vondrick, A.~Khosla, T.~Malisiewicz, and A.~Torralba.
\newblock {HOG}gles: Visualizing object detection features.
\newblock In {\em Proc. {ICCV}}, 2013.

\bibitem{wang10locality-constrained}
J.~Wang, J.~Yang, K.~Yu, F.~Lv, T.~Huang, and Y.~Gong.
\newblock Locality-constrained linear coding for image classification.
\newblock {\em Proc. {CVPR}}, 2010.

\bibitem{yang10supervised}
J.~Yang, K.~Yu, and T.~Huang.
\newblock Supervised translation-invariant sparse coding.
\newblock In {\em Proc. {CVPR}}, 2010.

\bibitem{ZeilerArxiv13}
M.~D. Zeiler and R.~Fergus.
\newblock Visualizing and understanding convolutional networks.
\newblock {\em CoRR}, abs/1311.2901, 2013.

\bibitem{zhou2014places}
B.~Zhou, A.~Lapedriza, J.~Xiao, A.~Torralba, and A.~Oliva.
\newblock {Learning Deep Features for Scene Recognition using Places Database.}
\newblock {\em NIPS}, 2014.

\bibitem{zhou10image}
X.~Zhou, K.~Yu, T.~Zhang, and T.~S. Huang.
\newblock Image classification using super-vector coding of local image
  descriptors.
\newblock In {\em Proc. {ECCV}}, 2010.

\end{thebibliography}
\end{document}